%% file: main.tex
\documentclass[11pt,a4paper]{article}
\usepackage{times,latexsym}
\usepackage{url}
\usepackage[T1]{fontenc}

\usepackage{graphicx}
\usepackage{graphics}
\usepackage{mdwlist}

\usepackage{algorithm}
\usepackage{algorithmic}

\usepackage{subfig}
\usepackage{multirow}

\usepackage{color}
\definecolor{org}{rgb}{0.992, 0.753, 0.525} 

\usepackage[acceptedWithA]{tacl2018v2}

\usepackage{xspace,mfirstuc,tabulary}

\newif\iftaclinstructions
\taclinstructionstrue 
\iftaclinstructions
\renewcommand{\confidential}{}
\newcommand{\instr}
\fi

\iftaclpubformat 

\else

\fi

\usepackage{flushend}

\title{Language Models with Transformers}
\author{
 Chenguang Wang \qquad Mu Li \qquad Alexander J. Smola \\
  Amazon Web Services \\
  {\sf \{chgwang, mli, smola\}@amazon.com} \\
}

\date{}

\begin{document}
\maketitle

\input{abstract}
\input{introduction}
\input{effectivelm} 
\input{search}
\input{experiments}
\input{relatedwork}
\input{conclusion}

\bibliography{tacl2018}
\bibliographystyle{acl_natbib}
\end{document}

%% file: abstract.tex
\begin{abstract}
    
The Transformer architecture is superior to RNN-based models in computational efficiency. Recently, GPT and BERT demonstrate the efficacy of Transformer models on various NLP tasks using pre-trained language models on large-scale corpora. Surprisingly, these Transformer architectures are suboptimal for language model itself. Neither self-attention nor the positional encoding in the Transformer is able to efficiently incorporate the word-level sequential context crucial to language modeling. 

In this paper, we explore effective Transformer architectures for language model, including adding additional LSTM layers to better capture the sequential context while still keeping the computation efficient. We propose Coordinate Architecture Search (CAS) to find an effective architecture through iterative refinement of the model. Experimental results on the PTB, WikiText-2, and WikiText-103 show that CAS achieves perplexities between $20.42$ and $34.11$ on all problems, i.e.\ on average an improvement of 12.0 perplexity units compared to state-of-the-art LSTMs. The source code is publicly available~\footnote{\url{https://github.com/cgraywang/gluon-nlp-1/tree/lmtransformer/scripts/language_model}}.

\end{abstract}

%% file: introduction.tex
\section{Introduction}

Modeling the sequential context in language is the key to success in many NLP tasks. Recurrent neural networks (RNNs)~\cite{mikolov2010recurrent} memorize the sequential context in carefully designed cells. The sequential nature of these models, however, makes computation expensive~\cite{Merity02182,Yang03953}, and therefore it is difficult to scale to large corpora.

The Transformer architecture~\cite{vaswani2017attention} replaces RNN cells with self-attention and point-wise fully connected layers, which are highly parallelizable and thus cheaper to compute. Together with positional encoding, Transformers are able to capture long-range dependencies with vague relative token positions. This results in a coarse-grained sequence representation at sentence level. Recent works such as GPT (or GPT-2)~\cite{radford2018improving,radford2019language} and BERT~\cite{devlin2018bert} show that the representations learned on large-scale language modeling datasets are effective for fine-tuning both sentence-level tasks, such as GLUE benchmark~\cite{wang2018glue}, and token-level tasks that do not rely on word order dependency in the context, such as question answering and NER.

Despite the fact that both GPT and BERT use language models for pre-training, neither of them achieves state-of-the-art performance in language modeling. Language model aims to predict the next word given the previous context, where fine-grained order information of words in context is required. Neither self-attention nor positional encoding in the existing Transformer architecture is effective in modeling such information. 

A second challenge (and opportunity) arises from the fact that we may often have access to models pre-trained on related, albeit not identical tasks. For instance, neither GPT or BERT is tuned for WikiText and neither of them aims to minimize perplexity directly. In fact, the architectures may not even be useful directly: BERT provides estimates of $p(w_i|\mathrm{context})$ rather than $p(w_i|\mathrm{history})$. This shows that there is a need for us to design algorithms which systematically explore the space of networks that can be derived (and adapted) from such tasks. This generalizes the problem of making use of pre-trained word embeddings for related tasks, only that in our case we do not have vectors but rather entire networks to deal with. 

Lastly, the problem of architecture search per-se has received great interests. However, the size of the datasets where training a single model for GPT or BERT can cost in excess of \$10,000, makes it prohibitively expensive to perform a fully-fledged model exploration with full retraining. Instead, we propose to use architecture search in a much more restricted (and economical) manner to investigate refining a trained architecture. This is much cheaper. Our pragmatic approach leads to improvements on the state-of-the-art in language modeling. 
Our contributions are as follows:
\begin{enumerate*}
    \item We propose a Transformer architecture for language model. It works by adding LSTM layers after all Transformer blocks (a result of the search algorithm). This captures fine-grained word-level sequential context.
    \item We describe an effective search procedure, Coordinate Architecture Search (CAS). This algorithm randomly generates variants of the Transformer architecture, based on the current best found architecture. Due to its greedy nature, CAS is simpler and faster than previous architecture search algorithms~\cite{ZophL16,PhamGZLD18,Liu09055}.
    \item We show how this can be used to incorporate substantial prior knowledge in the form of GPT or BERT. Using this information via brute force architecture search would be prohibitively expensive.
\end{enumerate*}
Contributions 2 and 3 are general and apply to many cases beyond NLP. Contribution 1 is arguably more language specific. We evaluate CAS on three popular language model datasets: PTB, WikiText-2 and WikiText-103. The BERT-based CAS achieves in average 12.0 perplexity gains compared to the state-of-the-art LSTM-based language model AWD-LSTM-MoS~\cite{Yang03953}.

%% file: effectivelm.tex
\section{Transformers for Language Models}
\label{sec:arch_for_language model}

Our Transformer architectures are based on GPT and BERT. We will reuse the pre-trained weights in GPT and BERT to fine-tune the language model task. During fine-tuning, we modify and retrain the weights and network used by GPT and BERT to adapt to language model task. 

\subsection{GPT and BERT}


{\bf GPT}~\cite{radford2018improving} uses a variant of the Transformer architecture~\cite{vaswani2017attention}. That is, it employs a multi-layer Transformer decoder based language model. The original paper provides a pre-trained architecture with 12-layer Transformer decoder-only blocks. Each block has hidden size 768 and 12 self-attention heads. The weights are trained  on BooksCorpus. This allows it to generate $p(w_i|\mathrm{history})$, one word at a time. 

{\bf BERT} is a multi-layer bidirectional Transformer encoder~\cite{devlin2018bert}. The original paper provides two BERT structures: {BERT-Base},  consists of 12-layer bidirectional Transformer encoder block with hidden size 768 and 12 self-attention heads; {BERT-Large} includes 24-layer bidirectional Transformer encoder blocks with hidden size 1024 and 16 self-attention heads. The weights are trained on BooksCorpus and the English Wikipedia. 
Unless stated otherwise, we mean {BERT Base} when mentioning {BERT}.

{\bf Relation between GPT and BERT.} Both models use virtually the same architecture. In fact, GPT and BERT-Base even use the same number of layers and dimensions. The only difference is that BERT is bidirectional since it tries to fill in individual words given their context, whereas GPT uses masked self-attention heads.

\subsection{Adapting GPT and BERT for Sub-word Language Model}
\label{sec:adapt}

GPT needs little modification, unless we want to explore different architectures. After all, it is already trained as a language model. At a minimum, during fine-tuning we add a linear layer with hidden size equal to the vocabulary size. These weights are tuned and fed into the softmax to generate a probability distribution of the target word over the vocabulary. Masked self-attention ensures that only causal information flow can occur. 

Recall the objective of BERT: masked language model and next sentence prediction. The masked language model uses bidirectional contextual information and randomly masks some tokens during training. Based on that it tries to infer the identity of the masked word. Unfortunately, estimating $p(w_i|w_1, \ldots w_{i-1}, w_{i+1}, \ldots w_n)$ is not conducive to building an effective text generator: We would need to design a Gibbs sampler to sample $w_i|w^{-i}$, i.e.\ $w_i$ given its context $w^{-i}$ iteratively and repeatedly for all $i$ to use a variant of this aspect directly. 

The next sentence prediction aims to capture the binarized relationship between two sentences. Again, this is not directly useful for LM. We thus remove the objective and replace it by a log-likelihood measure during fine-tuning. Similar to GPT, we add an output linear layer and replace the self-attention heads with masked self-attention to prevent leftward information flow.

Note that GPT and BERT pre-trained weights are re-used in the language model fine-tuning process to save the costs of a full retraining. We are thus conducting the language model in the sub-word level since the sub-word tokenization is used in both GPT and BERT. More details will be described in Section~\ref{sec:exp}. 

\subsection{Fine-tuning Transformer Weights}

GPT and BERT tune the weights of their respective models for the tasks
mentioned above. For instance, BERT doesn't use windowing by
default. Hence it makes sense to adjust the weights when fine-tuning
for language modeling. However, updating all weights could lead to
overfitting since datasets such as WikiText or Penn Tree Bank are over
an order of magnitude smaller than the data used to train GPT and
BERT.

To address this dilemma we propose to update only a subset of layer
weights during fine-tuning. Since both GPT and BERT have 12
Transformer blocks, each of which contains a self-attention and a
point-wise fully connected layer, it is not straightforward to choose
the subset of layers whose parameters should be fixed. Instead, we
will automatically search the subset which is most effective for the
language model task. The search algorithm will be discussed in
Section~\ref{sec:algorithm}.

\subsection{Adding an LSTM}

The positional encoding via a Fourier base in the Transformer only provides vague relative position information, forcing the layers to reinvent trigonometry at \emph{each} layer for specific word access. This is problematic since LM requires  strong word-level context information to predict the next word. RNNs explicitly model this sequential information. We therefore propose to add LSTM layers to the Transformer architecture. 

In theory we could add LSTM layers anywhere, even interleaving them with Transformers. However, LSTMs add significant computational efficiency penalties, since they prevent parallel computation. Our reasoning is analogous to that guiding the design of the SRU (simple recurrent unit) \cite{lei2018simple}. Hence we propose to add an LSTM layer either before all basic Transformer blocks or after these blocks. For the former, we add the LSTM layer immediately after the embedding layer and remove the positional and segment embedding, because we believe the LSTM layer is able to encode sufficient sequential information. For the latter, we insert the LSTM layer between the last Transformer block and the output linear layer. We determine the best location for the LSTM by automatic search. 

%% file: search.tex
\section{Coordinate Architecture Search}
\label{sec:derive}
\label{sec:algorithm}

Now that we have the basic components, let's review the network transformations and the associated search procedures to obtain a well-performing architecture.  

\begin{figure*}[t!]
    \centering
    \includegraphics[width=.77\textwidth]{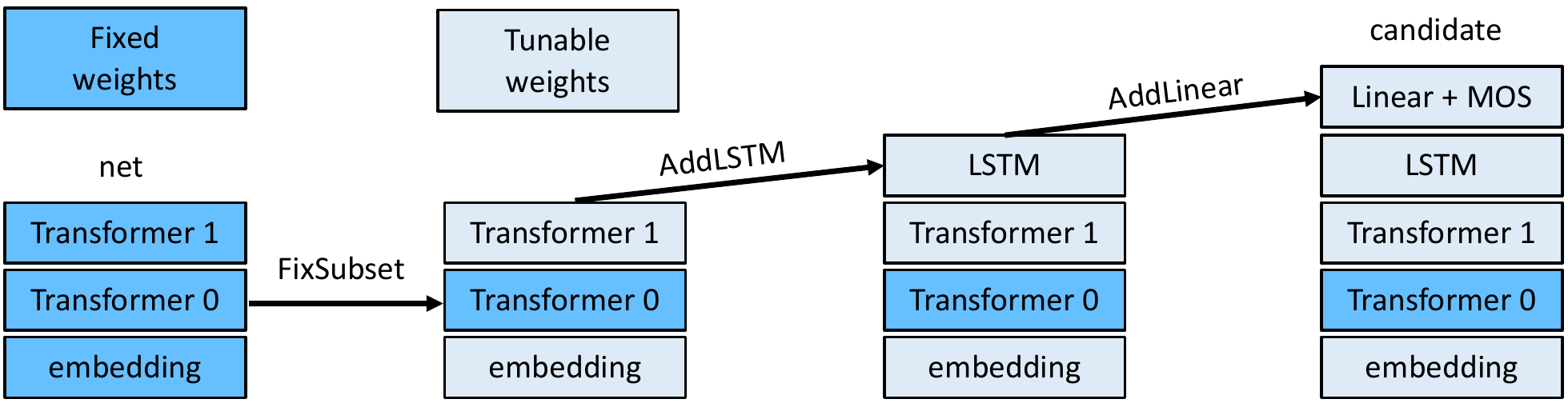}
    \caption{Search candidate sampling. \emph{net} is the base
      architecture and \emph{candidate} is returned in the next
      step. Transformers, Embeddings, LSTMs
      and Linear output transformations are as
      stated. Lightly shaded blocks are variable, dark blocks are fixed. See Algorithm~\ref{alg:sample} for details. 
      \label{fig:model-sample}}
    \smallskip
    \includegraphics[width=.9\textwidth]{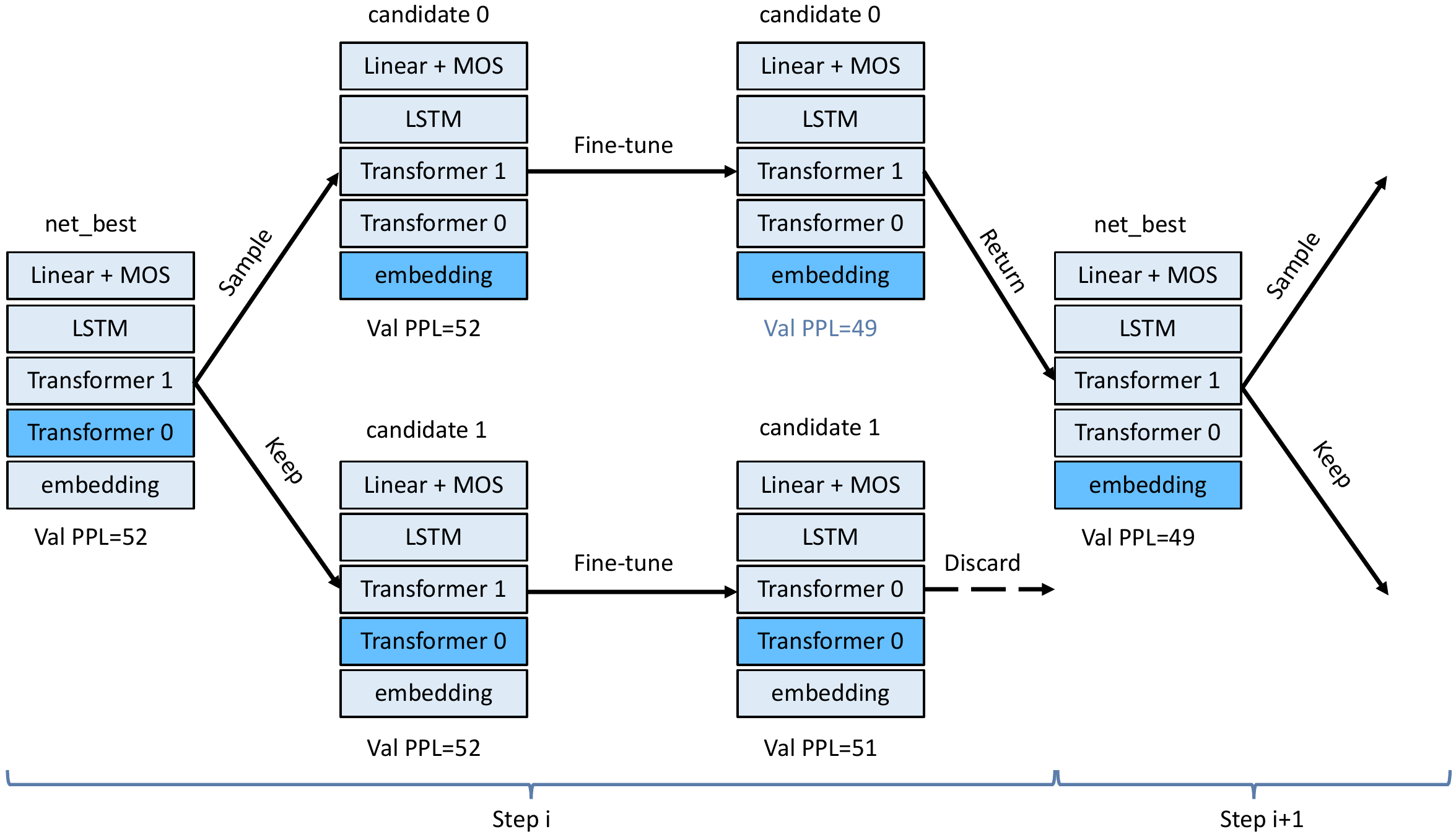}
    \caption{Coordinate architecture search. \emph{net\_best} is the best
      architecture at step $i$ of the search. We sample search
      candidates and keep the one that performs best, as measured by
      perplexity (Val PPL) on the target
      dataset after fine-tuning. See details
      in Algorithm~\ref{alg:search}. 
      \label{fig:searchlanguage}}
\end{figure*}

\subsection{Network Transformations}
\label{sec:nettransformation}

Transformations modify a network. A modification could be adding a new layer or fixing the parameters during fine-tuning. In Section~\ref{sec:arch_for_language model}, we proposed multiple transformations. Let us define them formally below with randomization and practical constraints.

\begin{description*}
\item[AddLinear] adds a linear output layer with hidden size equal to the vocabulary size. It then randomly initializes its parameters. If such a linear layer already exists, this step is skipped.
\item[AddLSTM] adds an LSTM layer if no such layer already exists. It attaches the LSTM either before or after all Transformer blocks. For the former we remove both positional embedding and segment embedding. If there exist fewer than 3 LSTM layers, we append another LSTM layer to the LSTM block. We randomly initialize parameters for the newly added layer.
\item[FixSubset] Given $n$ Transformer blocks, pick $k \in [0, n]$ uniformly at random. Accordingly pick $k$ blocks uniformly at random among the $\{1, \ldots n\}$ layers and fix the parameters for each selected block during fine-tuning.
\end{description*}

\subsection{Sampling a Search Candidate}

We need to generate architecture candidates during search. To
illustrate that restricted search is competitive to a
full-fledged brute force reinforcement learning (or genetic algorithms)
search, we adopt an exceedingly simple procedure: uniform random
sampling. At each time we sample transformations uniformly at random
(as per Algorithm~\ref{alg:sample}) from the set of modifications of a
base architecture until termination, as indicated by adding an
emissions layer $\mathit{AddLinear}$. This means that we have a valid
architecture. See Figure~\ref{fig:model-sample} for an example. 

\begin{algorithm}[tb]
\caption{\label{alg:sample}{Search Candidate Sampling}}
\begin{algorithmic}[1]
\REQUIRE{Base architecture \emph{net}}
\ENSURE{A new architecture \emph{candidate}}
\STATE \emph{candidate} $\gets$ \emph{net}
\REPEAT
\STATE Sample a tranformation $\mathcal{T}$ uniformly from $\{\mathit{AddLinear}, \mathit{AddLSTM}, \mathit{FixSubset}\}$.
\STATE Apply $\mathcal{T}$ to \emph{candidate}
\UNTIL{$\mathcal{T} = \mathit{AddLinear}$}
\RETURN \emph{candidate}
\end{algorithmic}
\end{algorithm}

\subsection{Coordinate Architecture Search}

We use a simple greedy strategy for architecture search. Starting with either
GPT or BERT as pre-trained model we repeat the search $n$ times. Each
time we sample a candidate, then fine-tune it and update the best
candidate if necessary. See Algorithm~\ref{alg:search} for a
description and Figure~\ref{fig:searchlanguage} for an
illustration. At each (successful) step we fine-tune the variable
parameters of the architecture. 

\begin{algorithm}[tb]
\caption{\label{alg:search}{Coordinate Architecture Search}}
\begin{algorithmic}[1]
\REQUIRE{Initial architecture \emph{net}, search steps $n$,
  fine-tuning dataset}
\ENSURE{Best architecture $\mathit{net}_\mathrm{best}$}
\STATE $\mathit{net}_\mathrm{best} \leftarrow \mathit{net}$;
\FOR{$i = 1$ {\bfseries to} $n$}
\STATE Draw $\mathit{candidate}$ from $\mathit{net}_\mathrm{best}$ using Algorithm~\ref{alg:sample}.
\STATE Fine-tune $candidate$ on dataset
\IF{$\mathrm{PPL}(\mathit{candidate}) < \mathrm{PPL}(\mathit{net}_\mathrm{best})$}
\STATE $\mathit{net}_\mathrm{best} \gets \mathit{candidate}$
\ENDIF
\ENDFOR
\RETURN $\mathit{net}_\mathrm{best}$
\end{algorithmic}
\end{algorithm}


%% file: experiments.tex
\section{Experiments}
\label{sec:exp}

To illustrate the effectiveness of the Transformer architectures found using coordinate search we present results on both WikiText and Penn TreeBank datasets. We also provide details about its speed relative to existing neural search strategies. 

\subsection{Datasets and Evaluation Metric}

We evaluate the proposed methods on three widely-used language model
benchmark datasets. {\bf Penn TreeBank (PTB)}: we use the preprocessed
version of \cite{mikolov2010recurrent}, which contains 100M
tokens. {\bf WikiText-2 (WT-2)} is a small preprocessed version of
Wikipedia, containing 200M tokens \cite{MerityXBS16}. {\bf
  WikiText-103 (WT-103)} contains 1B tokens of the same origin as
WT-2. We use the commonly adopted training, validation and test
splits.

To illustrate the flexibility of our approach, we explore two
pre-trained Transformers for sub-word level language models, i.e., BERT and GPT.  For
BERT related model architectures, we use WordPiece
embedding \cite{wu2016google} to tokenize the
training/validation/test split of the PTB, WT-2 and WT-103
respectively. The resulting sub-word vocabulary size is 30k, denoted as
{\it BERTVocab}. The split word pieces are denoted with $\#\#$
following \cite{devlin2018bert}.  For the model architectures based on
GPT, the three datasets are tokenized based on bytepair encoding
(BPE) \cite{SennrichHB16a}, where the sub-word vocabulary size is 40k
based on \cite{radford2018improving}, denoted as {\it GPTVocab}.  Note
that BERT and the WordPiece embedding in BERT are trained on
BooksCorpus and Wikipedia, whereas GPT and its BPE are trained only on
BooksCorpus. 
Note the sub-word level vocabulary size is different from the word-level vocabulary size obtained on the training splits of the datasets. 
We use perplexity (PPL) to evaluate the sub-word language model results.

\subsection{Training Details}

We evaluate CAS~(Algorithm~\ref{alg:search}) with both BERT and GPT pre-trained as the initial architecture, and trained on all three datasets. 
The same training configuration is used across all datasets.
We pick $n=10$ search steps. In a fine-tuning task, the number of epochs is 50, the gradients are computed using truncated back-propagation through time, and ADAM \cite{kingma2014adam} is used to update parameters. The perplexity on the validation dataset is used to choose architectures. We report results on the respective test datasets.

For GPT based architectures the hyperparameters of the Transformer decoder and embedding blocks are the same as in \cite{radford2018improving}. If LSTM layers are added, we set the dropouts of the LSTM layers to 0.1. DropConnect is not applied. All other LSTM hyperparameters follow \cite{Merity02182}. The final linear layer is with dropout rate 0.1. Following \cite{Yang03953}, we use a mixture of softmax (MoS) to replace the standard softmax with 15 components. We set 64 as sequence length and 16 as minibatch size. ADAM with learning rate $6.25 \cdot 10^{-5}$ and L2 weight decay of $0.01$ are used.

For BERT based architectures the hyperparameters of the Transformer encoder blocks and the embedding blocks are set the same as the original implementation \cite{devlin2018bert}. The hyperparameters of the LSTM layers and linear layer are the same with GPT configuration. As with GPT we use MoS with 15 components. We pick 128 as sequence length and 16 as minibatch size. ADAM with learning rate $10^{-4}$, $\beta_1 = 0.9$, $\beta_2 = 0.999$ and L2 weight decay of $0.01$ are used.

Lastly, for AWD-LSTM-MoS with BERT or GPT sub-word setting, we largely follow the parameter settings in the original implementation \cite{Yang03953}. We use NT-ASGD \cite{Merity02182} to train 50 epochs on training datasets.

Since the goal of this work is to discover best-performing language model from the architecture perspective, we do not employ post-training methods such as neural cache model \cite{GraveJU16} or dynamic evaluation \cite{KrauseK0R18}. We expect that such methods would potentially improve the perplexity of all models.

\begin{table*}[t!]
\centering
\begin{tabular}{|l|l|l|l|l|l|l|}
\hline
\multicolumn{1}{|c|}{\multirow{3}{*}{\textbf{Model}}} & \multicolumn{6}{c|}{\textbf{Datasets}}                                                                                                                                                                                   \\ \cline{2-7} 
\multicolumn{1}{|c|}{}                                & \multicolumn{2}{c|}{\textbf{PTB}}                                      & \multicolumn{2}{c|}{\textbf{WT-2}}                                     & \multicolumn{2}{c|}{\textbf{WT-103}}                                   \\ \cline{2-7} 
\multicolumn{1}{|c|}{}                                & \multicolumn{1}{c|}{\textbf{Val}} & \multicolumn{1}{c|}{\textbf{Test}} & \multicolumn{1}{c|}{\textbf{Val}} & \multicolumn{1}{c|}{\textbf{Test}} & \multicolumn{1}{c|}{\textbf{Val}} & \multicolumn{1}{c|}{\textbf{Test}} \\ \hline
AWD-LSTM-MoS-BERTVocab                                & 43.47                             & 38.04                              & 48.48                             & 42.25                              & 54.94                             & 52.91                              \\ \hline
BERT                                      & 72.99                             & 62.40                              & 79.76                             & 69.32                              & 109.54                            & 107.30                             \\ \hline
BERT-CAS (Our)                          & 39.97                    & 34.47                     & 38.43                    & 34.64                     & 40.70                   & 39.85                     \\ \hline 
BERT-Large-CAS (Our)                          & \textbf{36.14}                    & \textbf{31.34}                     & \textbf{37.79}                    & \textbf{34.11}                     & \textbf{19.67}                    & \textbf{20.42}                     \\ \hline
\hline
AWD-LSTM-MoS-GPTVocab                                 & 50.20                             & 44.92                              & 55.03                             & 49.77                              & 52.90                             & 51.88                              \\ \hline
GPT                                       & 79.44                             & 68.79                              & 89.96                             & 80.60                              & 63.07                             & 63.47                              \\ \hline
GPT-CAS (Our)                           & \textbf{46.24}                    & \textbf{40.87}                     & \textbf{50.41}                    & \textbf{46.62}                     & \textbf{35.75}                    & \textbf{34.24}                         \\ \hline
\end{tabular}
\caption{Performance of Coordinate Architecture Search (CAS). `Val' and `Test' denote validation and test perplexity respectively.}
\label{tab:result}
\end{table*}

\subsection{Comparing CAS to Other Methods}

We compare CAS, denoted by \textbf{BERT-CAS} and \textbf{GPT-CAS} respectively to three other models. 

{\bfseries BERT and GPT.} This is straightforward. The only change needed is that we update the last output layer during fine-tuning. 

{\bfseries AWD-LSTM-MoS-\{BERT, GPT\}Vocab.}  This is a state-of-the-art
language model, based on LSTMs, improving on \cite{Yang03953} due to a
more careful handling of tokens. For a fair comparison, instead of using word level vocabulary in the original implementation of AWD-LSTM-MoS~\cite{Yang03953}, we use the sub-word vocabularies of BERT and GPT separately. Our implementation uses
BERTVocab or GPTVocab to replace the word based vocabulary used in
the original implementation of AWD-LSTM-MoS \cite{Yang03953}. Note
that on PTB and WT-2, both AWD-LSTM-MoS-BERTVocab and
AWD-LSTM-MoS-GPTVocab outperform the original AWS-LSTM-MoS models by
17.8 and 10.6 perplexity points respectively. This is likely due to
the change in word vocabulary to a sub-word vocabulary.

\begin{figure*}[tb]
  \includegraphics[width=\textwidth]{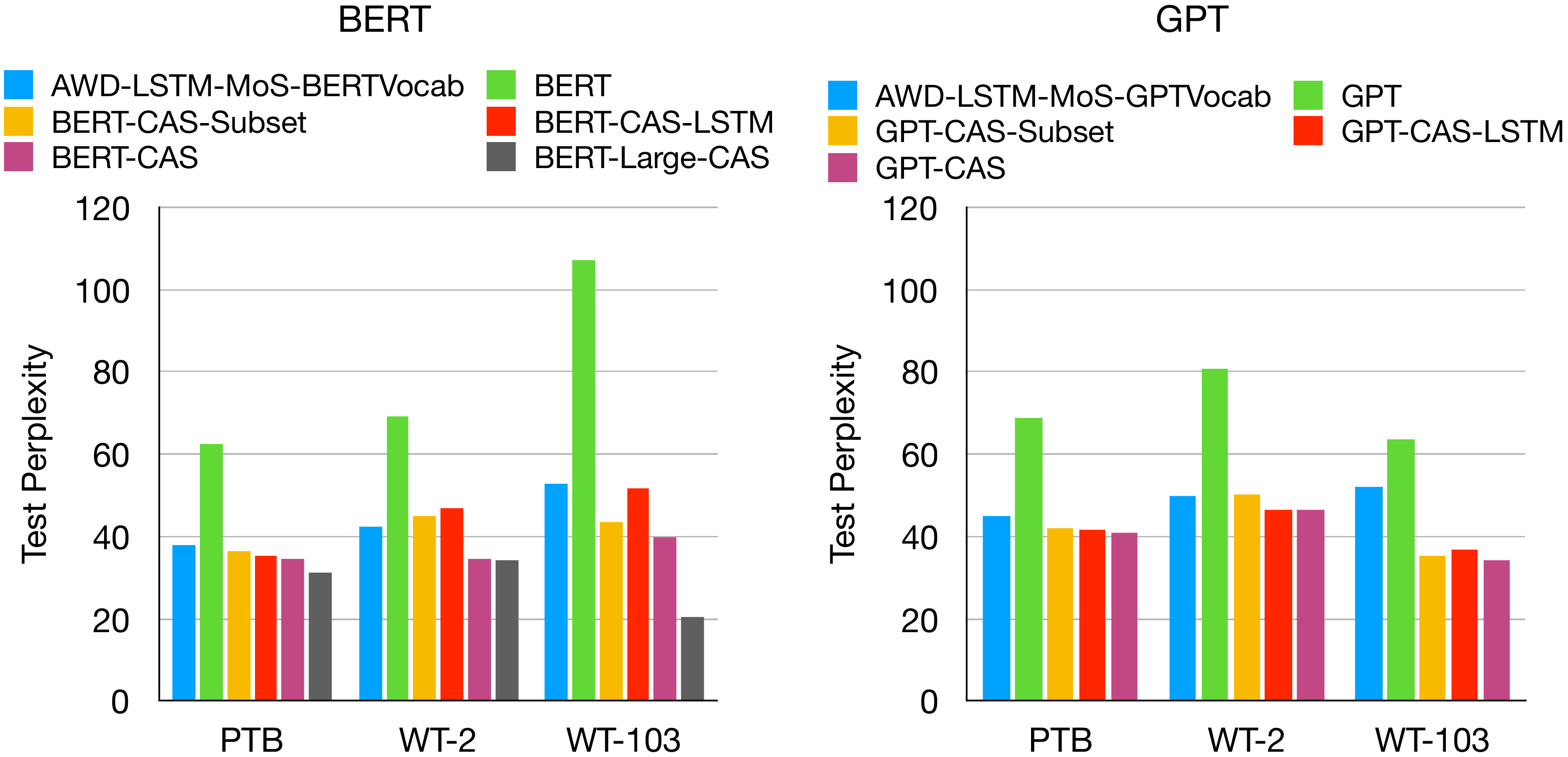}
\caption{Comparison of test perplexities between CAS and other
  models (left: using BERT pre-trained models; right: using GPT
  pre-trained models). In particular, `Subset' indicates variants
  without LSTMs and `LSTM' corresponds to models without updating the
  transformer blocks. 
  \label{fig:ppl}}
\end{figure*}

The results are shown in Table~\ref{tab:result} and illustrated in
Figure~\ref{fig:ppl}. First note that GPT and BERT are significantly
worse than AWD-LSTM-MoS. It confirms our hypothesis that neither BERT
nor GPT are effective tools for language modeling. Applying them
naively leads to significantly worse results compared to AWS-LSTM-MoS
on three datasets. It demonstrates that language modeling requires
strong capabilities in modeling the word order dependency within
sentences. However, due to the combination of self-attention and
positional encoding, both GPT and BERT primarily capture
coarse-grained language representation but with limited word-level
context.

On the other hand, the Transformer architectures picked by CAS (BERT-CAS)
outperform AWS-LSTM-MoS on all datasets. The average test perplexity
improvement with BERT pre-trained models is 8.09 and 8.28 with GPT
pre-trained models. The results demonstrate that 1) CAS is able to
locate an effective Transformer architecture for language model; and
2) that the combination of fixing a subset weights and adding LSTM
layers is capable of capturing the word-level context.

Furthermore, we apply CAS to BERT-Large (i.e., BERT-Large-CAS). Compare to BERT-CAS, the architectures generated achieve on average 7.70 perplexity gains, which are competitive results with recent approaches such as Transformer-XL~\cite{dai2019transformer} and GPT-2~\cite{radford2019language}. This shows that robustness of the CAS method, which indicates that a stronger pre-trained model would potentially produce a better language model.

In addition, BERT-CAS outperforms GPT-CAS on datasets PTB and WT-2, but is worse on WT-103. The reason is twofold. First, the GPT's BPE vocabulary is 10k larger than BERT's WordPiece vocabulary, since the original word vocabulary size of WT-103 is around 10 times larger compared to PTB and WT-2, thus we infer that BPE vocabulary has stronger ability to represent large vocabulary. Second, unlike GPT, the pre-trained BERT weights are not based on a language modeling objective. Thus BERT based architectures may need more epochs to converge on large corpora. This is likely due to the fact that masking is not a part of BERT training. Its introduction amounts to a more significant change in the covariates, thus requires more adaptation.




\begin{table*}[t!]
\centering
\begin{tabular}{|l|l|l|l|l|l|l|}
\hline
\multicolumn{1}{|c|}{\multirow{3}{*}{\textbf{Model}}} & \multicolumn{6}{c|}{\textbf{Datasets}}                                                                                                                                                                                   \\ \cline{2-7} 
\multicolumn{1}{|c|}{}                                & \multicolumn{2}{c|}{\textbf{PTB}}                                      & \multicolumn{2}{c|}{\textbf{WT-2}}                                     & \multicolumn{2}{c|}{\textbf{WT-103}}                                   \\ \cline{2-7} 
\multicolumn{1}{|c|}{}                                & \multicolumn{1}{c|}{\textbf{Val}} & \multicolumn{1}{c|}{\textbf{Test}} & \multicolumn{1}{c|}{\textbf{Val}} & \multicolumn{1}{c|}{\textbf{Test}} & \multicolumn{1}{c|}{\textbf{Val}} & \multicolumn{1}{c|}{\textbf{Test}} \\ \hline
BERT-CAS-Subset                                   & 42.53                             & 36.57                              & \textbf{51.15}                             & \textbf{44.96}                              & \textbf{44.34}                             & \textbf{43.33}                              \\ \hline
BERT-CAS-LSTM               & \textbf{40.22}                             & \textbf{35.32}                              & 53.82                             & 47.00                              & 53.66                             & 51.60                              \\ \hline\hline
GPT-CAS-Subset                                    & 47.58                             & 41.85                              & 54.58                             & 50.08                              & \textbf{35.49}                             & \textbf{35.48}                              \\ \hline
GPT-CAS-LSTM                & \textbf{47.24}                             & \textbf{41.61}                              & \textbf{50.55}                             & \textbf{46.62}                     & 36.68                             & 36.61                              \\ \hline

\end{tabular}
\caption{Ablation study. Compare CAS with not adding LSTM layers (CAS-Subset) and not updating Transformer block parameters (CAS-LSTM). }
\label{tab:abl}
\end{table*}

\subsection{Ablation Study}

To elucidate the effects of different model improvements we compare
CAS to the following three variants:
\begin{description*}
\item[\{BERT, GPT\}-CAS-Subset] applies Algorithm~\ref{alg:search} without adding LSTM layers.
\item[\{BERT, GPT\}-CAS-LSTM] applies Algorithm~\ref{alg:search} but
  it fixes all Transformer blocks during fine-tuning.
\end{description*}
See Table~\ref{tab:abl} and Figure~\ref{fig:ppl} for details of the
results. As can be seen, both CAS-Subset and CAS-LSTM improve
significantly upon a naive use of BERT and GPT. This is to be expected
since fine-tuning improves performance. On the smaller dataset, i.e.\
PTB, adding LSTMs is more effective. This might be due to the
overfitting incurred in updating Transformers. On the other hand, on
the larger datasets, i.e.\ WT-103, adding an LSTM is less effective,
which means that adapting the Transformer parameters for better
sentence-level representation is more important. Combing both together
leads to further improvement. CAS outperforms AWD-LSTM-MoS on all
three datasets.

Next, we unfreeze the pre-trained weights of BERT to allow fully fine-tuning including the last linear output layer (BERT-All) on PTB data as an example, to illustrate the over-fitting issue. From the results in Table~\ref{tab:overfit}, we can see that, by leveraging CAS, we marginally relieve the over-fitting issue by fixing a subset of weights of the full Transformer architecture.

\begin{table}[]
\centering
\begin{tabular}{|l|l|l|}
\hline
\textbf{Model} & \textbf{Validation} & \textbf{Test} \\ \hline
BERT-All      & 79.14               & 67.43         \\ \hline
BERT-CAS       & \textbf{39.97}               & \textbf{34.47}         \\ \hline
\end{tabular}
\caption{Over-fitting example on PTB data. BERT-All: BERT with fully fine-tuning including the last layer. BERT-CAS: BERT with coordinate architecture search.}
\label{tab:overfit}
\end{table}

\begin{figure}[tb]
    \centering
    \includegraphics[width=\columnwidth]{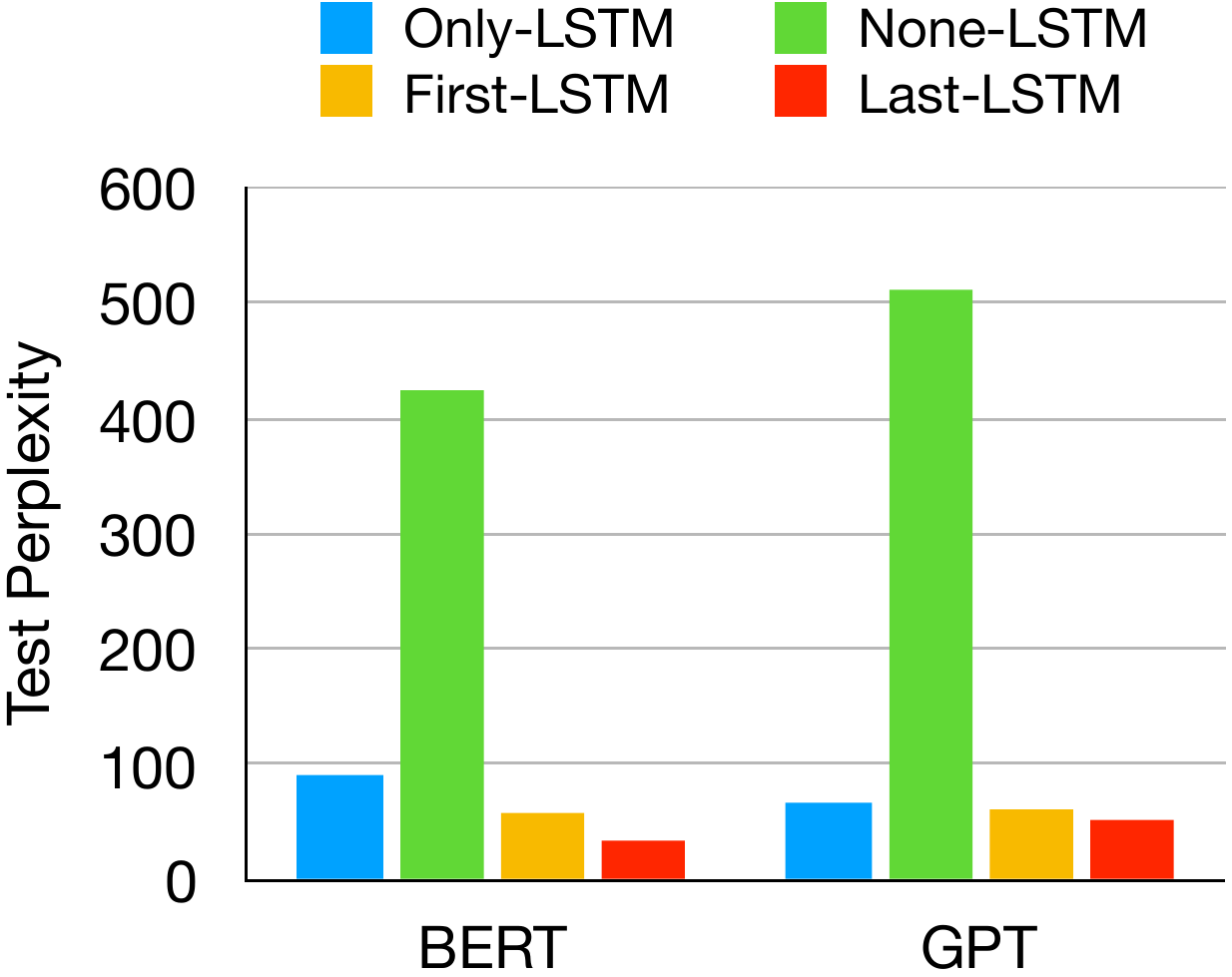}
    \caption{LSTM variants for Penn TreeBank. We study whether to add
      LSTMs before or after the transformer layers (or none at all).
      \label{fig:lstm}}
\end{figure}

Let's look into the details of adding LSTMs. There are 4 cases:
\begin{description*}
\item [Only-LSTM] implements a model consisting only of LSTM
  layers. To make up for the loss of expressiveness due to removing
  all Transformer blocks we add a total of 6 LSTM layers. 
\item [None-LSTM] adds no LSTM layer at all. Instead, we add another
  stack of Transformer blocks. This effectively doubles the number of
  blocks to 24.
\item [First-LSTM] adds LSTM layers only before all Transformer blocks. 
\item [Last-LSTM] adds LSTM layers only after all Transformer blocks. 
\end{description*}
The results are shown in Table~\ref{tab:effect} and in
Figure~\ref{fig:lstm}. As can be seen, neither purely transformer
blocks nor purely LSTM layers are effective for language modeling. The
former is likely unsuitable due to the comparatively large number of
parameters relative to the tuning set. Adding LSTM layers properly
into Transformer architecture significantly improves the
perplexity. In addition, adding LSTMs before the output linear layer
outperforms replacing positional and segment embeddings with LSTM
layers. These results confirm our intuition and indicate that we need
to first preserve the coarse-grained representation using fixed subset
weights; subsequently LSTMs can be used to model the word order
dependency.

\begin{table}[tb]
{\centering
\begin{tabular}{|lll|r|r|}
\hline
\multicolumn{3}{|c|}{\textbf{Constraints}} & \textbf{Validation} & \textbf{Test} \\ \hline
\multirow{4}{*}{BERT} 
& Only & \multirow{4}{*}{LSTM}                       & 107.47                            & 89.82                              \\ 
& None &                        & 491.75                            & 425.32                             \\ 
& First&                       & 67.85                             & 57.71                              \\ 
& Last&                        & {\bf 39.97}                             & {\bf 34.47}                              \\ \hline 
\multirow{4}{*}{GPT} 
& Only & \multirow{4}{*}{LSTM}                       & 75.76                             & 66.56                              \\ 
& None &                         & 579.77                            & 510.49                             \\ 
& First &                        & 70.05                             & 60.82                              \\ 
& Last &                         & {\bf 46.24}                             & {\bf 40.87}                              \\ \hline
\end{tabular}
\caption{Effects of different search constraints for placing the LSTM
  on perplexity on the PTB data. }
\label{tab:effect}
}
\end{table}

\subsection{Efficiency Analysis}

Lastly, we compare CAS with other existing neural network search
methods in terms of search cost. The main distinction being that we
significantly constrain the architectures to be investigated. This
allows us to obtain significant computational savings. 
\begin{description*}
\item[NAS] by \cite{ZophL16} is a reinforcement learning based search method, which uses a recurrent network to generate the model descriptions of neural networks and minimizes the expected perplexity of the generated architectures on the PTB validation set.
\item[ENAS] by \cite{PhamGZLD18} also leverages a reinforcement learning search method. ENAS's search space is the superposition of all possible child models in the NAS search space, which allows parameters to be shared among all child models.
\item[DARTS] by \cite{Liu09055} is a recently proposed neural architecture search algorithm based on gradient descent. 
\end{description*}
We evaluate the efficiency of the methods using GPU days. The search costs of NAS, ENAS and DARTS are obtained from  \cite{Liu09055}.

\begin{table*}[]
{\centering
\begin{tabular}{|l|l|l|l|}
\hline
\multicolumn{1}{|c|}{\multirow{2}{*}{\textbf{Search Method}}} & \multicolumn{2}{c|}{\textbf{\begin{tabular}[c]{@{}c@{}}Search Cost\\ (GPU days)\end{tabular}}} & \multicolumn{1}{c|}{\multirow{2}{*}{\textbf{Method Class}}} \\ \cline{2-3}
\multicolumn{1}{|c|}{}                                        & \multicolumn{1}{c|}{\textbf{PTB}}             & \multicolumn{1}{c|}{\textbf{WT-2}}             & \multicolumn{1}{c|}{}                                       \\ \hline
NAS \cite{ZophL16}                                                           & 1,000 CPU days                                  & n.a.                                              & reinforcement                                               \\ \hline
ENAS \cite{PhamGZLD18}                                                          & 0.5                                           & 0.5                                            & reinforcement                                               \\ \hline
DARTS (first order) \cite{Liu09055}                                           & 0.5                                           & \multirow{2}{*}{1}                             & gradient descent                                            \\ \cline{1-2} \cline{4-4} 
DARTS (second order) \cite{Liu09055}                                          & 1                                             &                                                & gradient descent                                            \\ \hline
BERT-CAS (Our)                           & \textbf{0.15}                                 & \textbf{0.38}                                  & greedy search                                               \\ \hline
GPT-CAS (Our)                            & 0.23                                 & 0.53                                           & greedy search                                               \\ \hline
\end{tabular}
\caption{Efficiency of different search methods on PTB and WT-2.}
\label{tab:eff}
}
\end{table*}

\begin{table*}[]
{
\centering
\begin{tabular}{|ll|l|l|l|l|l|l|}
\hline
\multicolumn{1}{|c|}{\multirow{3}{*}{\textbf{Model}}} & \multirow{3}{*}{\textbf{Parameters}} & \multicolumn{6}{c|}{\textbf{Datasets}}                                                                                                \\ \cline{3-8} 
\multicolumn{1}{|c|}{}                                &                                      & \multicolumn{2}{c|}{\multirow{2}{*}{PTB}} & \multicolumn{2}{c|}{\multirow{2}{*}{WT-2}} & \multicolumn{2}{c|}{\multirow{2}{*}{WT-103}} \\
\multicolumn{1}{|c|}{}                                &                                      & \multicolumn{2}{c|}{}                     & \multicolumn{2}{c|}{}                      & \multicolumn{2}{c|}{}                        \\ \hline
\multicolumn{1}{|l|}{\multirow{3}{*}{GPT-2}}          & 345M                                 & \multicolumn{2}{l|}{47.33}                & \multicolumn{2}{l|}{22.76}                 & \multicolumn{2}{l|}{26.37}                   \\ \cline{2-8} 
\multicolumn{1}{|l|}{}                                & 762M                                 & \multicolumn{2}{l|}{40.31}                & \multicolumn{2}{l|}{19.93}                 & \multicolumn{2}{l|}{22.05}                   \\ \cline{2-8} 
\multicolumn{1}{|l|}{}                                & 1542M                                & \multicolumn{2}{l|}{35.76}                & \multicolumn{2}{l|}{\textbf{18.34}}        & \multicolumn{2}{l|}{\textbf{17.48}}          \\ \hline
\multicolumn{1}{|l|}{BERT-Large-CAS}            & 395M                                 & \multicolumn{2}{l|}{\textbf{31.34}}       & \multicolumn{2}{l|}{34.11}                 & \multicolumn{2}{l|}{20.42}                   \\ \hline
\end{tabular}
\caption{Compare model parameter size and results with GPT-2. The GPT-2 model size and results are from~\cite{radford2019language}.}
\label{tab:compgpt2}
}
\end{table*}

\begin{table}[]
\centering
\begin{tabular}{|l|l|l|}
\hline
\multicolumn{1}{|c|}{\multirow{3}{*}{\textbf{Model}}} & \multirow{3}{*}{\textbf{Training Data}} & \multirow{3}{*}{\textbf{Tokens}} \\
\multicolumn{1}{|c|}{}                                &                                         &                                  \\
\multicolumn{1}{|c|}{}                                &                                         &                                  \\ \hline
GPT-2                                                 & WebText                                 & 14.0B                              \\ \hline
\multirow{3}{*}{BERT-Large-CAS}                 & PTB                                     & {\bf 0.1B}                             \\ \cline{2-3} 
                                                      & WT-2                                    & {\bf 0.2B}                             \\ \cline{2-3} 
                                                      & WT-103                                  & {\bf 1.0B}                           \\ \hline
\end{tabular}
\caption{Compare training data size with GPT-2.}
\label{tab:compgpt2data}
\end{table}

The reported search costs of the above methods compared to CAS are shown in Table~\ref{tab:eff}. As can be seen, BERT-CAS is cheaper than all others. The results indicate that by leveraging the prior knowledge of the design of the neural networks for specific tasks, we could only optimize the architectures in a small confined sub-space, that leads to speed up the search process. 
For example, BERT-CAS is directly based on BERT, applying search upon such effective neural networks could facilitate the adaptation to similar tasks.

The reason of the search cost of GPT-CAS on WT-2 is higher than ENAS
is three-fold: 
\begin{enumerate*}
\item ENAS is directly transferring an architecture searched based on PTB
to WT-2. Instead we apply coordinate search to find one from scratch; 
\item The model size of GPT-CAS is 149M, which is much larger compared to the size 37M from ENAS; 
\item The GPT vocabulary size is 10k larger compared to the ENAS's
  vocabulary.
\end{enumerate*}
We note that the difference in vocabularies might affect the results,
since the results of NAS, ENAS and DARTS are from the original
implementations. The original implementations are based on basic word tokenization (such as space splitter) of the PTB and WT-2. Instead, we
are using the sub-word tokenization (WordPiece and BPE respectively)
for BERT and GPT architecture exploration. However, the vocabulary
size after basic tokenization processing is similar to the results
after the sub-word tokenization, which are all around 30k-40k. Given
that, we consider the performance comparison as fair~\footnote{The PPL results are not comparable since the vocabularies are different (i.e., sub-word versus word level), we omit the comparison here.}.

\subsection{Comparison with GPT-2}

We specifically compare the proposed model with the recent state-of-the-art language model GPT-2~\cite{radford2019language} on three dimensions: results\footnote{The results comparison is fair since GPT-2's vocabulary is also based on sub-word tokenization.}, parameter size, and scale of the training data. From the results shown in Table~\ref{tab:compgpt2}, we conclude that with comparable size of the models' parameters, BERT-Large-CAS outperforms GPT-2 (345M) by on average {10.97} PPL on PTB and WT-103. More surprisingly, the proposed method performs better than GPT-2 (1542M) which has around 4 times more parameters. On WT-103, BERT-Large-CAS is better than GPT-2 (762M) which has around 2 times more parameters. Note that on WT-2, our method performs worse than GPT-2, we suspect the reason is that the WebText still contains the texts that are similar to the Wikipedia. WT-2 is quite small in terms of scale. In contrast, we regard the results on WT-103 (50 times larger than WT-2) as a more reasonable comparison with GPT-2. 

The training data described in Table~\ref{tab:compgpt2data} suggests that, with significantly smaller training datasets, the proposed method generates competitive results. Once GPT-2 models are released, we expect CAS could generalize to the GPT-2 models to obtain better results for language model task.

%% file: relatedwork.tex
\section{Related Work}

{\bf Architecture search} has shown promising results in tasks such as image classification \cite{ZophL16,liu2017progressive,liu2017hierarchical,real2018regularized,ZophVSL18,Liu09055}, object detection \cite{ZophVSL18} as well as language modeling \cite{ZophL16,PhamGZLD18,Liu09055} in NLP. Existing neural architecture search studies focus on leveraging different methods to build the neural network from scratch. For example, NAS \cite{ZophL16} uses reinforcement learning to obtain an architecture for CIFAR-10 and ImageNet. 
Designing the architecture from scratch using  reinforcement learning is very costly. Many follow-up studies focus on speeding up the search process by weight-sharing across child models \cite{PhamGZLD18,cai2018efficient}, by incorporating a particular structure into the search space \cite{liu2017progressive,liu2017hierarchical}, or by enabling weights prediction for each architecture \cite{brock2017smash,baker2017accelerating}. 
Different from the above methods, the proposed coordinate search does not involve any controllers. 

Recent studies start to explore using the idea of network transformation within reinforcement learning \cite{cai2018efficient} or via Bayesian optimization \cite{jin2018efficient} or simple greedy search \cite{Elsken04528}. DARTS \cite{Liu09055} enables gradient descent to optimize the architecture. 
Compared to these methods, the coordinate search is more straightforward and more efficient due to the direct incorporation of the pre-defined Transformer architecture. Notably, the major difference of the proposed search algorithm compared to the existing methods is that we focus on \emph{adapting} an existing well-trained Transformer architecture with minimum changes in the task of language model, whereas a majority of the existing work focus on generating variants of RNN cells \emph{from scratch} for better results.

{\bf Language models} have been studied extensively in NLP. Neural language models have supplanted traditional n-gram models in recent years 
\cite{bengio2003neural,mnih2007three,mikolov2010recurrent}. Particularly, recurrent neural networks \cite{inan2016tying,Merity02182,melis2017state,KrauseK0R18}, such as LSTMs have achieved state-of-the-art results on various benchmark datasets with different regularization techniques and post-training methods \cite{GraveJU16,KrauseK0R18}. 
The mixture of softmax \cite{Yang03953} has helped address the low-rank embedding problem for word prediction. We used this in our model, too. It provides some improvement over a more conventional model. 

The recently proposed GPT-2~\cite{radford2019language} is a deeper Transformer decoder based language model trained on a 40GB dataset. In contrast, the proposed model generates competitive results but with significantly less training cost and smaller model size. Transformer-XL~\cite{dai2019transformer} is a word level language model that also delivers good results by incorporating longer context. The proposed method is a sub-word level language model thus the results are not comparable. We expect to generalize CAS to pre-trained Transformer-XL models as well to achieve better results. The adaptive input representations idea proposed in~\cite{baevski2018adaptive} could be combined with the proposed method to further speed up.

{\bf Network transformations} were introduced in the context of the transfer learning \cite{ChenGS15}. The main purpose of the transformations is to make networks deeper and wider. Often stagewise training accelerates training and architecture search. Recent studies \cite{wei2016network,cai2018efficient,Elsken04528} focus on extending the set of the network transformations to handle additional operations such as non-linear activation functions and skip connections. We instead introduce simple network modifications to perform modest modifications of an existing network. They allow us to treat a pre-trained Transformer block in a manner similar to that of a large pre-trained embedding vector. 

%% file: conclusion.tex
\section{Conclusion}

We study the problem of finding an effective Transformer architecture for language model. 
We identify the issues of existing Transformer architectures, such as BERT and GPT, that are not able to capture the strong word-level context required in language model.
We proposed two approaches to address this issue: we fine-tune a subset of parameters to improve the coarse-grain representations obtained from the pre-trained Transformer models. Secondly, we add LSTM layers to capture the fine-grained sequence. We then propose a coordinate architecture search (CAS) algorithm to select an effective architecture based on fine-tuning results. It uses a greedy search strategy to accelerate architecture search. We experimentally show that CAS outperforms the state-of-the-art language models on three language model benchmark datasets.

Although we only show the effectiveness of CAS when applying Transformer architectures to the language model task, we feel it is possible to apply CAS to both other neural network architectures and fine-tuning other NLP tasks that require strong word-level context as well.